\documentclass[10pt,twocolumn,letterpaper]{article}

\usepackage[pagenumbers]{cvpr} 
\usepackage{algpseudocode}
\usepackage{algorithm}
\usepackage{multirow}
\usepackage{cuted}
\usepackage{colortbl}
\usepackage{threeparttable}
\usepackage[dvipsnames]{xcolor}
\usepackage[accsupp]{axessibility} 
\definecolor{cvprblue}{rgb}{0.21,0.49,0.74}
\usepackage[pagebackref,breaklinks,colorlinks,citecolor=cvprblue]{hyperref}

\def\paperID{2417} 
\def\confName{CVPR}
\def\confYear{2024}

\title{GeoAuxNet: Towards Universal 3D Representation Learning for Multi-sensor Point Clouds}

\author{Shengjun Zhang$^{1}$, Xin Fei$^{2}$, Yueqi Duan$^{1\dag}$\\
$^{1}$ Department of Electronic Engineering, Tsinghua University\\
$^{2}$ Department of Automation, Tsinghua University \\
{\tt\small \{zhangsj23, feix21\}@mails.tsinghua.edu.cn, duanyueqi@tsinghua.edu.cn}
}

\begin{document}

\maketitle
\newcommand\blfootnote[1]{%
\begingroup 
\renewcommand\thefootnote{}\footnote{#1}%
\addtocounter{footnote}{-1}%
\endgroup 
}
\blfootnote{\textsuperscript{\dag}Corresponding author.}
\begin{strip}
    \centering
    \vspace{-2.5cm}
    \begin{center}
    \textbf{\url{https://github.com/zhangshengjun2019/GeoAuxNet}}
    \end{center}
    \includegraphics[width=\linewidth]{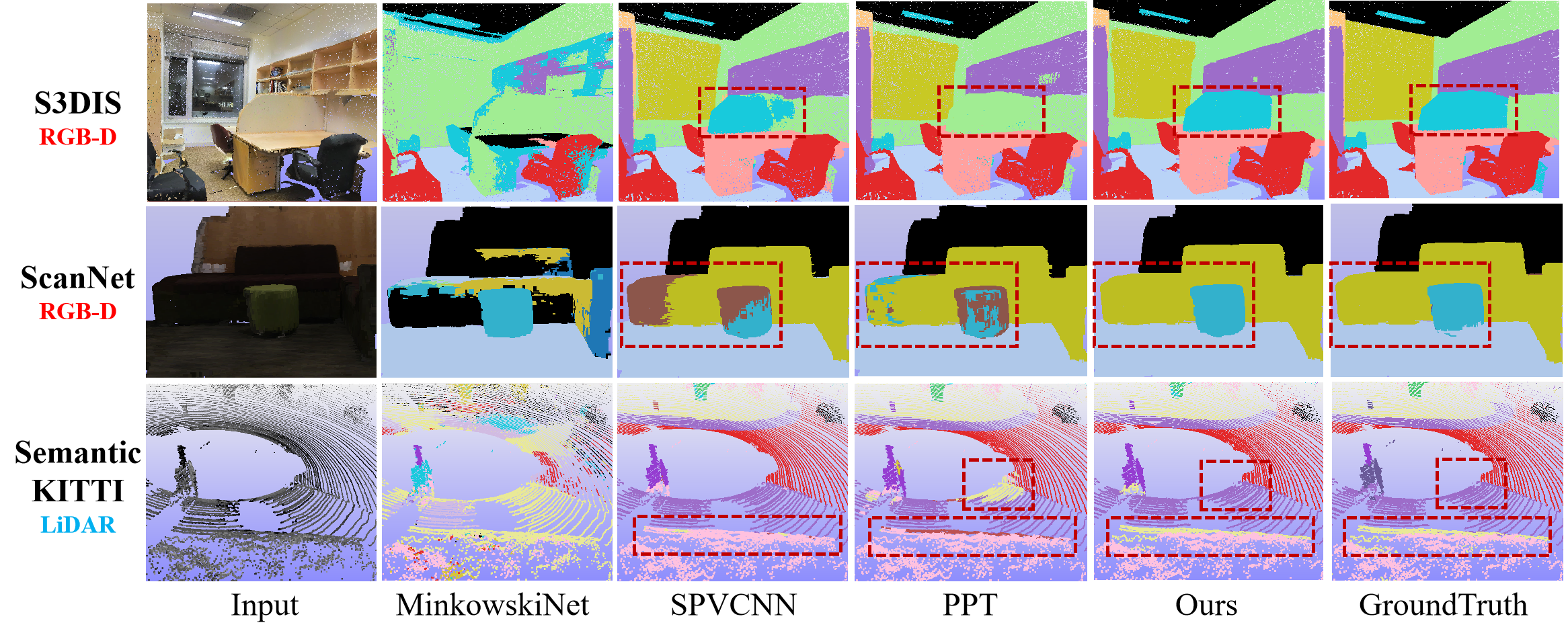}
    \captionof{figure}{Semantic segmentation results on S3DIS~\cite{S3DIS2016CVPR} and ScanNet~\cite{ScanNet2017CVPR} from RGB-D cameras and SemanticKITTI~\cite{SemanticKITTI2019ICCV} from LiDAR. For all methods, we trained collectively on three datasets. Our method outperforms other methods with better detailed structures.}
    \label{fig:teaser}
\end{strip}

\begin{abstract}
Point clouds captured by different sensors such as RGB-D cameras and LiDAR possess non-negligible domain gaps. 
Most existing methods design different network architectures and train separately on point clouds from various sensors. Typically, point-based methods achieve outstanding performances on even-distributed dense point clouds from RGB-D cameras, while voxel-based methods are more efficient for large-range sparse LiDAR point clouds. 
In this paper, we propose geometry-to-voxel auxiliary learning to enable voxel representations to access point-level geometric information, which supports better generalisation of the voxel-based backbone with additional interpretations of multi-sensor point clouds.
Specifically, we construct hierarchical geometry pools generated by a voxel-guided dynamic point network, which efficiently provide auxiliary fine-grained geometric information adapted to different stages of voxel features.
We conduct experiments on joint multi-sensor datasets to demonstrate the effectiveness of GeoAuxNet. Enjoying elaborate geometric information, our method outperforms other models collectively trained on multi-sensor datasets, and achieve competitive results with the-state-of-art experts on each single dataset.

\end{abstract}     
\section{Introduction}
\label{sec:intro}


Point cloud analysis has attracted widespread attention due to its growing applications, such as autonomous driving~\cite{DeepFusion2022CVPR, SparseFusion2023ICCV, Center3D2021CVPR} and robotics~\cite{MMAE2016RSS, ExplicitOR2011CVPR}.  
Unlike images which are represented by regular pixels, 3D point clouds are irregular and unordered.
These characteristics aggravate the inconsistency on density and sampling patterns of point clouds captured by different sensors, such as RGB-D cameras~\cite{S3DIS2016CVPR, ScanNet2017CVPR, SUNRGB-D2015CVPR} and LiDAR~\cite{SemanticKITTI2019ICCV, nuScene2019CVPR, Waymo2019CVPR}.

Typically, points generated from RGB-D pictures are equally distributed and dense, while points scanned by LiDAR are sparse and uneven. The diversity on input data hinders the construction of universal network architectures.
For point clouds from RGB-D cameras, point-based methods~\cite{PointNet++2017NIPS, PointTransformer2021ICCV, PointNeXt2022NIPS} usually gather local information via grouping and aggregating neighborhood features to extract detailed spatial information. Since neighbors are not stored contiguously in point representations, indexing them requires the costly searching operations~\cite{PVCNN2019NIPS}, which limits the application of these methods on large scale LiDAR point clouds.
To address this challenge, voxel-based methods~\cite{SPCNN2018CVPR, VoxNet2015IROS, OctNet2017CVPR} leverage the regular 3D convolution to obtain the contiguous memory access pattern. However, the resolution of voxels is constrained by the memory and becomes lower at deeper stages, resulting in loss of local information~\cite{SPVCNN2020ECCV}. 
An intuitive idea is to ensemble both point-based and voxel-based methods~\cite{PVCNN2019NIPS, PVT2021IJIS, SPVCNN2020ECCV}, yet they are still confronted with either high time-consumption or the lack of detailed geometric features, which fail to simultaneously process large scale point clouds from different sensors with fine-grained spatial information while maintaining efficiency. 

In this paper, we propose GeoAuxNet to provide point-level geometric information for voxel representations in an efficient manner with geometry-to-voxel auxiliary learning. The support of elaborate geometric features introduces additional point-level information, which cannot be fully exploited by voxel-based backbones. Motivated by the observation that the local geometric structures present high similarity at each stage of the network, we construct hierarchical geometry pools to learn auxiliary sensor-aware point-level geometric priors corresponding to different layers for sensor-agnostic voxel features. To update our geometry pools, we present a voxel-guided dynamic point network, where we leverage prior knowledge in voxel features to guide high quality spatial feature extraction. Then, we fuse elaborate geometric features in the pools into voxel representations via our designed geometry-to-voxel auxiliary mechanism. For large-scale point clouds, the hierarchical geometry pools store representative point-level features at different stages, which efficiently provide complementary geometric information without using the point-based networks during inference time.

We conduct extensive experiments on multi-sensor datasets, including S3DIS~\cite{S3DIS2016CVPR} and ScanNet~\cite{ScanNet2017CVPR} from RGB-D cameras as well as SemanticKITTI~\cite{SemanticKITTI2019ICCV} captured by LiDAR, to demonstrate the efficiency and effectiveness of our method. With a shared weight of backbone, our method outperforms other models trained on joint datasets from different sensors and achieves competitive performance with experts on single dataset. As shown in Figure~\ref{fig:teaser}, we trained MinkowskiNet~\cite{SparseUNet2019CVPR}, SPVCNN~\cite{SPVCNN2020ECCV}, PPT~\cite{PPT2023ArXiv} and GeoAuxNet on the above three datasets collectively, where our method preserve better detailed structures for point clouds from various sensors.

\begin{figure*}
    \centering
    \includegraphics[width=\linewidth]{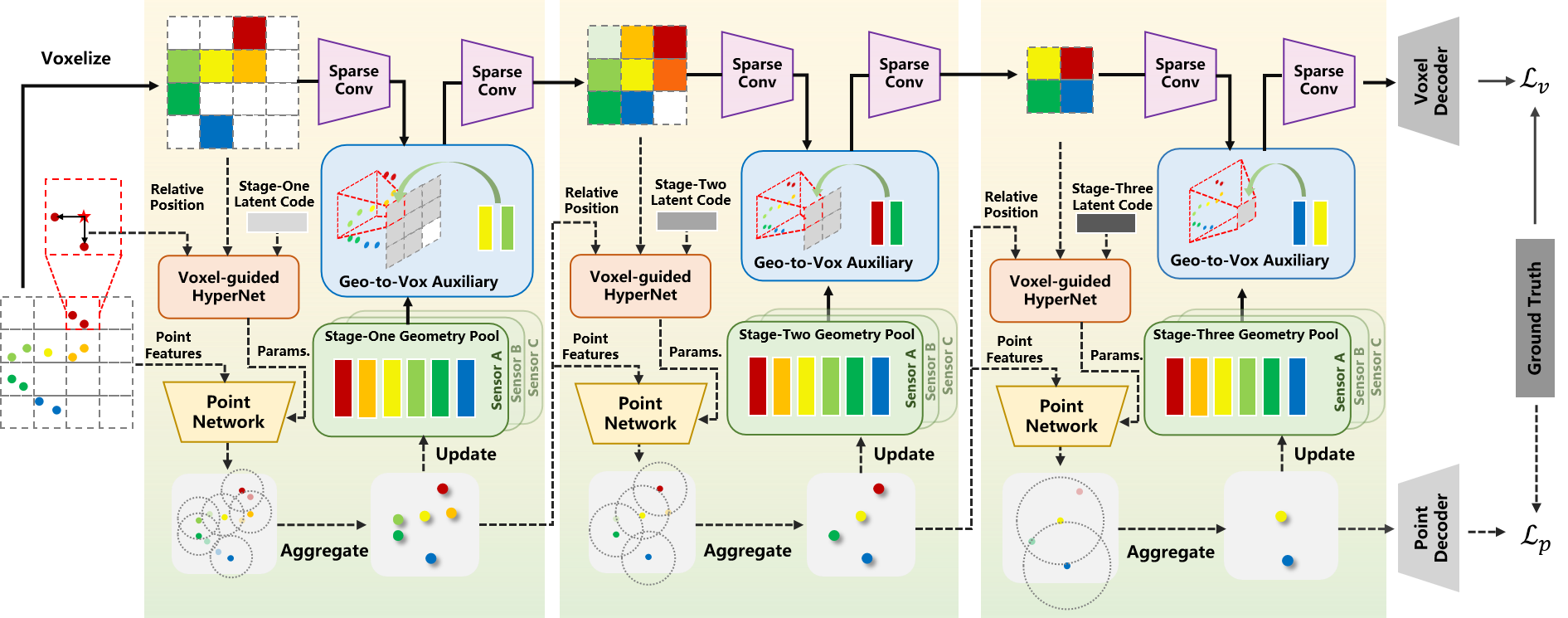}
    \caption{The pipeline of our GeoAuxNet. For a complete scene point cloud $\mathcal{P}^{\mathcal{C}}$ and a local point patch $\mathcal{P}\subseteq\mathcal{P}^{\mathcal{C}}$, our voxel-based backbone first voxelizes $\mathcal{P}^{\mathcal{C}}$ and conducts sparse convolutional operations. The voxel-guided hypernetwork takes relative positions, voxel features and a stage latent code as input to provide weights and biases for the point network. Then, we encode the spatial information for $\mathcal{P}$ with the point network and aggregate local features to generate geometric feature candidates. Following the update strategy, we construct hierarchical geometry pools. The geometry-to-voxel mechanism fuses geometric features stored in the pools to  enable voxel representations to access point-level geometric information. We repeat the above process several times to extract effective representation hierarchically and predict the results with a voxel decoder for the primary task and a point decoder for the auxiliary task. The dotted line stands for the course of the auxiliary learning which is ignored during inference to ensure efficiency. Geo-to-Vox is abbreviation of Geometry-to-Voxel.}
    \label{fig:pipeline}
    \vspace{-0.5cm}
\end{figure*}

\section{Related Works}
\label{sec:related works}

\paragraph{3D Scene Understanding.}
Since images are composed of regular pixels, point clouds generated from RGB-D cameras are well-distributed and dense. Voxelization on dense point clouds ignore detailed structure information. To directly learn from points, Qi~\textit{et al.}~\cite{PointNet2017CVPR} first proposed PointNet to process point clouds through shared multi-layer perceptions (MLP) to get a global representation. Yet this method lacks the ability to comprehend local geometric features. Therefore, PointNet++~\cite{PointNet++2017NIPS} built a hierarchical architecture to gather local information via grouping and aggregating neighborhood features. Following up MLP-based works~\cite{PointNeXt2022NIPS, PointMiXer2022ECCV} focused on designing effective learning schemes to capture spatial features.
Other methods also introduced specific convolution operations~\cite{PointCNN2018NIPS, ConvPoint2019CGF, KPConv2019ICCV, PAConv2021CVPR}, graph-based representations~\cite{3DGCN2020CVPR, DGCNN2018TOG} and self-attention mechanism~\cite{PointTransformer2021ICCV, PointTransformerV22022NIPS, PointTransformer2020IEEE, StratifiedTransformer2022CVPR, Swin3D2023} to explore the relationship between points.
Apart from cameras, LiDAR is also a widely used sensor to collect point clouds. 
Some methods~\cite{RangeFormer2023ArXiv, SqueezeSeg2018ICRA, PolarNet2020CVPR, RangeNet++2019IROS} focused on projecting point clouds to 2D grids to utilize 2D CNN. However, 3D to 2D projection limits the representation quality of geometric structures~\cite{STD2019ICCV} due to the sacrifice of 3D spatial information~\cite{PAConv2021CVPR}.
Point based methods have high time consumption because of the large scale and the sparsity of these point clouds. Voxel-based methods~\cite{Cylinder3D2021CVPR, SphereFormer2023CVPR, OccuSeg2020CVPR, SPCNN2018CVPR, 3DUNet2016MICCAI} show more robustness against the sparsity. In view of the sampling characteristic of LiDAR, Zhu~\textit{et al.}~\cite{Cylinder3D2021CVPR} introduced cylinder coordinate system, for nearby regions have greater density than farther away regions. Similarly, Lai~\textit{et al.}~\cite{SphereFormer2023CVPR} proposed spherical transformer to aggregate information from dense close points to sparse distant ones. 
Voxel-based networks maintain efficiency when processing scene-level point clouds, but ignore some detailed spatial structures. 

\paragraph{Universal Learning from Multiple Data Sources.}
Large language models achieve remarkable progress on many natural language processing systems through pretraining on extremely large text corpora~\cite{CommonCrawl2021EMNLP}, which enables foundation models to process data across domains, involving multiple languages, various majors, diverse application scenarios and so on. In 2D computer vision, joint learning across domains improve model robustness for detection~\cite{Universal-RCNN2020AAAI, DA2019CVPR, SimpleMDD2021CVPR}, depth estimation~\cite{MonoDepth2019TPAMI} and semantic segmentation~\cite{MSeg2021TPAMI, CDCL2021AAAI}. Wang~\textit{et al.}~\cite{DA2019CVPR} proposed a universal object detector through a domain-attention mechanism. But they did not model the training differences between different datasets. Universal-RCNN~\cite{Universal-RCNN2020AAAI} modeled the class relations by designing a inter-dataset attention module to incorporates graph transfer learning for propagating relevant semantic information across multiple datasets. CDCL~\cite{CDCL2021AAAI} pointed out that convolution filters can be shared across domains without accuracy loss, while normalization operators are not appropriate to share in view of the bias of statistics. Recently, Wu~\textit{et al.}~\cite{PPT2023ArXiv} introduced point prompt training to optimize a learnable domain-specific prompt with language-guided categorical alignment, but processed point clouds from RGB-D and LiDAR separately. However, point clouds captured by different sensors have diverse density and distributions, which limits the exploration of the universal network on multiple data sources.

\paragraph{Auxiliary Learning.}
Auxiliary learning aims to achieve outstanding performance on a single primary task with the assist of auxiliary tasks and has shown many successful applications in a wide range of fields, including knowledge distillation~\cite{KDwithSS2020ECCV}, transfer learning~\cite{TransferAuxiliary2015ICCV} and reinforcement learning~\cite{RLAuxiliary2019NIPS}. For instance, DCAN~\cite{DCAN2016CVPR} learned an auxiliary contour detection task for more accurate segmentation, while Araslanov investigated the joint learning of image classification and semantic segmentation. In 3D vision, MonoCon~\cite{MonoCon2021AAAI} introduced monocular contexts as auxiliary tasks in training for 3D object detection. Nevertheless, the inherent relationship between point-level features and voxel-level features is still under exploration to better leverage the advantages of two domains without any requirement of additional data.

\section{Method}

The overall framework of GeoAuxNet is illustrated in Figure~\ref{fig:pipeline}. To generate point-level geometric features, we propose a voxel-guided hypernetwork to produce weights and biases for the point network. Then, point features are grouped and aggregated to extract fine-grained local features, which are employed to update geometry pools. For voxel-based backbone, we introduce geometry-to-voxel auxiliary mechanism to access elaborate spatial information in our hierarchical geometry pools. 

\subsection{Voxel-guided Dynamic Point Network}\label{sec:voxel-guide extraction}

Point-based methods are mostly confronted with high time-consumption towards large scale point clouds, since the farthest point sampling (FPS) and k-nearest neighbors (kNN) algorithms in these networks have a time complexity of $\mathcal{O}(n^{2})$.
Therefore, to efficiently process large scale point clouds, input points are voxelized and passed through sparse convolution operations to extract voxel features in our voxel-based backbone. 

To generate point-level geometric information efficiently, we train our point network on local point patches instead of complete scene point clouds. However, less training data weakens the ability of the point network to extract local features. Considering that our voxel-based branch are trained on complete voxelized point clouds, we treat voxel features as prior knowledge to instruct the learning of local geometric information. Yet, points inside a voxel share the same voxel feature, which ignores detailed difference between points. Thus, we also introduce relative coordinates to depict diversity of points within the same voxel. On account of the hierarchical architectures of both the point network and the voxel-based backbone with several stages, we optimize a learnable latent code $s$ for each stage during our auxiliary learning.

Given a local point patch $\mathcal{P}=\{p_{i}, 1\leq i\leq N\}\in \mathbb{R}^{N\times 3}$ with point features $\mathcal{F}^{\mathcal{P}}=\{f^{\mathcal{P}}_{i}, 1\leq i\leq N\}$, each point $p_{i}$ belongs to a voxel $v$ with the voxel feature $f^{\mathcal{V}}$.
Here $v\in\mathbb{R}^{3}  $ stands for the center coordinate of the voxel.
Then, we can obtain the weight and bias for point $p_{i}$:
\begin{align}
    w_{i} & =h_{w}(h_{p}(p_{i}-v),f^{\mathcal{V}})\odot h_{s}(s), \\
    b_{i} & =h_{b}(h_{p}(p_{i}-v),f^{\mathcal{V}})\odot h_{s}(s), 
\end{align}
where $h_{p}$ and $h_{s}$ are MLPs to project relative positions and the stage latent code to specific dimensions, and $h_{w}$ and $h_{b}$ are hypernetworks to generate the weight and bias.
Hence, new point features can be formulated as:
\begin{equation}
    \hat{f}^{\mathcal{P}}_{i}=\text{MLP}(w_{i}\odot f^{\mathcal{P}}_{i}+b_{i}).
\end{equation}
To extract local geometric information, we use kNN algorithm to select $k$ nearest neighbor points and employ a max-pooling function to aggregate each local group. In this manner, we generate a set of geometric feature candidates for hierarchical geometry pools. 

\subsection{Hierarchical Geometry Pools}
The voxel branch ignores some detailed spatial structures due to voxelization. Motivated by the similarity of geometric structures in 3D space, we establish a pool to store typical geometric patterns. The pool contains $n$ typical spatial structure features $\mathcal{F}^{\mathcal{D}}=\{f^{\mathcal{D}}_{i}, 1\leq i\leq n\}$. The max size of the pool is set to $N$. In Section~\ref{sec:voxel-guide extraction}, we generate a set of geometric pattern candidates $\mathcal{F}^{\mathcal{S}}=\{f^{\mathcal{S}}_{i}, 1\leq i\leq m\}$. When updating the pool, we first calculate the similarity between $\mathcal{F}^{\mathcal{S}}$ and $\mathcal{F}^{\mathcal{D}}$:
\begin{align}
    s_{i} = \mathop{max}\limits_{1\leq j\leq N} s_{i,j}, \label{eq:cos-sim deputation}
\end{align}
where
\begin{equation}
    s_{i,j} = \dfrac{f^{\mathcal{S}}_{i}\cdot f^{\mathcal{D}}_{j}}{\Vert f^{\mathcal{S}}_{i}\Vert \Vert f^{\mathcal{D}}_{j}\Vert}.
\end{equation}
Specifically, if $s_{i}\geq \epsilon$,  $f^{\mathcal{S}}_{i}$ is merged into the geometry pool $\mathcal{F}^{\mathcal{D}}$:
\begin{align}
    t & =\mathop{argmax}\limits_{1\leq j\leq N} s_{i,j}, \\
    \hat{f}^{\mathcal{D}}_{t} & =\lambda f^{\mathcal{D}}_{t}+(1-\lambda)f^{\mathcal{S}}_{i},\label{eq:absorb}
\end{align}
where $\epsilon$ is a threshold and $\lambda$ stands for the update rate. 
Otherwise, we append the new features to the geometry pool. If the total number of features is more than $N$, we randomly select $N-n$ new features and add them to the pool. The rest new features following Eq.~\ref{eq:absorb} are merged into existing patterns. Our update strategy is illustrated in Algorithm~\ref{algorithm}.

Since voxel representations at various stages contains different level information, we introduce hierarchical geometry pools to store representative point-level features corresponding to different stages. We also enlarge the size of pools for deeper stages due to more complex geometric structures with larger receptive field sizes.

\begin{algorithm}[t]
\caption{Update Pools}\label{algorithm}
\renewcommand{\algorithmicrequire}{\textbf{Input:}}
\renewcommand{\algorithmicensure}{\textbf{Output:}}
\begin{algorithmic}
\Require New features $\mathcal{F}^{\mathcal{S}}$, geometry pool $\mathcal{F}^{\mathcal{D}}$, maximum size of the pool $N$, threshold $\epsilon$, update rate $\lambda$
\Ensure New pool $\hat{\mathcal{F}}^{\mathcal{D}}$
\State Compute $s_{i}$ for $1\leq i\leq m$ according to Eq.~\ref{eq:cos-sim deputation}
\For{$i\leftarrow 1$ to $\vert\mathcal{F}^{\mathcal{S}}\vert$}
    \If{$s_{i}\geq \epsilon$}
        \State $t\leftarrow\arg\max_{1\leq j\leq N}{s_{i,j}}$
        \State $f^{\mathcal{D}}_{t}\leftarrow\lambda f^{\mathcal{D}}_{t}+(1-\lambda)f^{\mathcal{S}}_{i}$
        \State $\mathcal{F}^{\mathcal{S}}\leftarrow\mathcal{F}^{\mathcal{S}}\setminus\{f^{\mathcal{S}}_{i}\}$
    \EndIf
\EndFor
\If{$\vert\mathcal{F}^{\mathcal{S}}\vert+\vert\mathcal{F}^{\mathcal{D}}\vert\leq N$}
    \State $\mathcal{F}^{\mathcal{D}}\leftarrow \mathcal{F}^{\mathcal{D}}\cup\mathcal{F}^{\mathcal{S}}$
\Else
    \State $\hat{\mathcal{F}}^{\mathcal{S}}\leftarrow$ randomly select $N-\vert\mathcal{F}^{\mathcal{D 
  }}\vert$ features from $\mathcal{F}^{\mathcal{S}}$
    \State $\mathcal{F}^{\mathcal{S}}\leftarrow\mathcal{F}^{\mathcal{S}}\setminus\hat{\mathcal{F}}^{\mathcal{S}}$
    \State $\mathcal{F}^{\mathcal{D}}\leftarrow \mathcal{F}^{\mathcal{D}}\cup\hat{\mathcal{F}}^{\mathcal{S}}$
    \For{$i\leftarrow 1$ to $\vert\mathcal{F}^{\mathcal{S}}\vert$}
        \State $t\leftarrow\arg\max_{1\leq j\leq N}{s_{i,j}}$
        \State $f^{\mathcal{D}}_{t}\leftarrow\lambda f^{\mathcal{D}}_{t}+(1-\lambda)f^{\mathcal{S}}_{i}$
    \EndFor
\EndIf
\end{algorithmic}
\end{algorithm}

\subsection{Geometry-to-Voxel Auxiliary}
After we update the geometric pool, we use geometric patterns stored in $\mathcal{F}^{\mathcal{D}}\in\mathbb{R}^{N\times C}$ to generate fine-grained local information for each voxel following a cross-attention operation. Given voxel features $\mathcal{F}^{\mathcal{V}}=\{f^{\mathcal{V}}_{i}, 1\leq i\leq M\}\in\mathbb{R}^{M\times D}$, the geometric fine-grained feature $f^{\mathcal{G}}_{i}$ for $f_{i}^{\mathcal{V}}\in \mathcal{F}^{\mathcal{V}}$ is formulated by:
\begin{align}
    q_{i}&=h_{Q}(f_{i}^{\mathcal{V}})\in \mathbb{R}^{1\times s}, \\
    K&=h_{K}(\mathcal{F}^{\mathcal{D}})\in \mathbb{R}^{N\times s},\\
    V&=h_{V}(\mathcal{F}^{\mathcal{D}})\in \mathbb{R}^{N\times t}, \\
    f^{\mathcal{G}}_{i}&=softmax\left(\dfrac{q_{i}K^\top}{\sqrt{s}}\right)V\in \mathbb{R}^{1\times t},
\end{align}
where $h_{Q}$, $h_{K}$, and $h_{V}$ stand for projections. 
Then, we concatenate $f^{\mathcal{G}}_{i}$ with $f^{\mathcal{V}}_{i}$ as the new voxel feature and employ sparse convolution operations. 
In this manner, we can provide complementary geometric information without using the point-based networks during inference stage.

\begin{table*}[t]
\centering
\begin{threeparttable}[b]
\caption{Semantic segmentation results on three benchmarks including S3DIS~\cite{S3DIS2016CVPR}, ScanNet~\cite{ScanNet2017CVPR} and SemanticKITTI~\cite{SemanticKITTI2019ICCV}. We train SPVCNN~\cite{SPVCNN2020ECCV}, PPT~\cite{PPT2023ArXiv} and GeoAuxNet on the joint training data of three datasets and also compare with experts~\cite{PointTransformer2021ICCV, StratifiedTransformer2022CVPR, PointNeXt2022NIPS, Cylinder3D2021CVPR, SphereFormer2023CVPR, SPVCNN2020ECCV} on each single dataset. We report the class-average accuracy (mAcc, \%) and class-wise mean IoU (mIoU, \%). The type stands for point-based (\textit{p}) or voxel-based (\textit{v}) methods, where \textit{p}$\rightarrow$\textit{v} means the geometry-to-voxel auxiliary learning.}
\begin{tabular}{lcc|@{\hspace{3mm}}llllll}
\toprule
\multirow{2}{*}{Method} & \multirow{2}{*}{Type} & \multirow{2}{*}{Params.} 
& \multicolumn{2}{c}{S3DIS Area5}       
& \multicolumn{2}{c}{ScanNet} 
& \multicolumn{2}{c}{SemanticKITTI} \\ \cline{4-9} 
\specialrule{0em}{3pt}{0pt} 
&
&
& Test mIoU      
& Test mAcc   
& Val mIoU     
& Val mAcc    
& Val mIoU        
& Val mAcc        \\ \midrule
\rule{0pt}{10pt} 
PointTransformer~\cite{PointTransformer2021ICCV} 
& \textit{p}
& 11.4M
& 70.4
& 76.5 
& 70.6      
& - 
& \cellcolor{gray!10}{}             
& \cellcolor{gray!10}{} \\
\rule{0pt}{10pt}
StratifiedFromer~\cite{StratifiedTransformer2022CVPR}  
& \textit{p}
& 18.8M   
& 72.0 
& 78.1    
& 74.3          
& -    
& \cellcolor{gray!10}{}               
& \cellcolor{gray!10}{} \\
\rule{0pt}{10pt}
PointNeXt~\cite{PointNeXt2022NIPS}     
& \textit{p}
& 41.6M   
& 70.5           
& 77.2    
& 71.5          
& -      
& \cellcolor{gray!10}{}              
& \cellcolor{gray!10}{} \\
\rule{0pt}{10pt}
SPVCNN~\cite{SPVCNN2020ECCV}  
& \textit{p}$+$\textit{v}
& 21.8M   
& \cellcolor{gray!10}{}              
& \cellcolor{gray!10}{}  
& \cellcolor{gray!10}{}        
& \cellcolor{gray!10}{} 
& 63.8            
& -  \\
\rule{0pt}{10pt}
Cylinder3D~\cite{Cylinder3D2021CVPR}   
& \textit{v}
& 26.1M   
& \cellcolor{gray!10}{}              
& \cellcolor{gray!10}{}  
& \cellcolor{gray!10}{}
& \cellcolor{gray!10}{}            
& 64.3            
& -   \\
\rule{0pt}{10pt}
SphereFormer~\cite{SphereFormer2023CVPR}  
& \textit{v}
& 32.3M   
& \cellcolor{gray!10}{}              
& \cellcolor{gray!10}{}
& \cellcolor{gray!10}{}
& \cellcolor{gray!10}{}            
& 67.8            
& -    \\ \midrule
\rule{0pt}{10pt}
MinkowskiNet~~\cite{SparseUNet2019CVPR}             
& \textit{v}
& 60.9M   
& \cellcolor{yellow!10}{}18.8          
& \cellcolor{yellow!10}{}21.3  
& \cellcolor{yellow!10}{}25.6          
& \cellcolor{yellow!10}{}30.5
& \cellcolor{yellow!10}{}36.0           
& \cellcolor{yellow!10}{}41.2                \\
\rule{0pt}{10pt}
SPVCNN~\cite{SPVCNN2020ECCV}          
& \textit{p}$+$\textit{v}
& 61.0M   
& \cellcolor{yellow!10}{}58.6           
& \cellcolor{yellow!10}{}62.3  
& \cellcolor{yellow!10}{}56.7          
& \cellcolor{yellow!10}{}60.4  
& \cellcolor{yellow!10}{}52.0            
& \cellcolor{yellow!10}{}56.7            \\
\rule{0pt}{10pt}
PPT~~\cite{PPT2023ArXiv}             
& \textit{v}
& 63.0M   
& \cellcolor{yellow!10}{}30.4          
& \cellcolor{yellow!10}{}34.6  
& \cellcolor{yellow!10}{}16.5          
& \cellcolor{yellow!10}{}27.4
& \cellcolor{yellow!10}{}25.0           
& \cellcolor{yellow!10}{}31.2                \\
\rule{0pt}{10pt}
PPT~\tnote{$\dagger$}~~\cite{PPT2023ArXiv}             
& \textit{v}
& 63.0M   
& \cellcolor{yellow!10}{}67.9          
& \cellcolor{yellow!10}{}73.2  
& \cellcolor{yellow!10}{}69.6          
& \cellcolor{yellow!10}{}77.5 
& \cellcolor{yellow!10}{}61.8           
& \cellcolor{yellow!10}{}68.9             \\
\rule{0pt}{10pt}
GeoAuxNet                 
& \textit{p}$\rightarrow$\textit{v}
& 64.7M   
& \cellcolor{yellow!10}{}$\textbf{69.5}_{\textcolor{Green}{(+1.6)}}$           
& \cellcolor{yellow!10}{}$\textbf{74.5}_{\textcolor{Green}{(+1.3)}}$   
& \cellcolor{yellow!10}{}$\textbf{71.3}_{\textcolor{Green}{(+1.7)}}$
& \cellcolor{yellow!10}{}$\textbf{79.3}_{\textcolor{Green}{(+1.8)}}$     
& \cellcolor{yellow!10}{}$\textbf{63.8}_{\textcolor{Green}{(+2.0)}}$ 
& \cellcolor{yellow!10}{}$\textbf{69.3}_{\textcolor{Green}{(+0.4)}}$               \\ \bottomrule
\end{tabular}
\label{tab:main results}
\begin{tablenotes}
    \small
     \item[$\dagger$] The original PPT with language-guided categorical alignment fails to converge on the joint training data. We use decoupled projection heads to employ PPT on multi-sensor datasets.
   \end{tablenotes}
\end{threeparttable}
\end{table*}

\subsection{Overall Architecture}
Given a point cloud $\mathcal{P}_{i}^{\mathcal{C}}$ with features $\mathcal{F}^{\mathcal{C}}_{i}$ and a local point patch $\mathcal{P}_{i}\subseteq\mathcal{P}_{i}^{\mathcal{C}}$ with $\mathcal{F}^{\mathcal{P}}_{i}\subseteq\mathcal{F}_{i}^{C}$ from sensor $\mathcal{S}_{i}$, we voxelize all points to obtain voxel coordinates $\mathcal{V}_{i}$ with voxel features $\mathcal{F}^{\mathcal{V}}_{i}$. 
Points obtained from RGB-D cameras usually possess colors and normal vectors, while points captured by LiDAR always possess intensity. Thus, we introduce sensor-aware input embedding. Specifically, we embed $\mathcal{P}_{i}$ from sensor $\mathcal{S}_{i}$ by:
\begin{align}
    \hat{\mathcal{F}}^{\mathcal{P}}_{i} & = h^{\mathcal{P}}_{\mathcal{S}_{i}}(\mathcal{F}^{P}_{i}), \\
    \hat{\mathcal{F}}^{\mathcal{V}}_{i} & = h^{\mathcal{V}}_{\mathcal{S}_{i}}(\mathcal{F}^{V}_{i}), 
\end{align}
where $h^{\mathcal{P}}_{\mathcal{S}_{i}}$ means shared MLPs on points and $h^{\mathcal{V}}_{\mathcal{S}_{i}}$ means sparse convolution operations on voxels.
Then, we learn voxel features and point features following the extraction strategy in Section~\ref{sec:voxel-guide extraction} and update hierarchical geometry pools using Algorithm~\ref{algorithm}.
We fuse point-level geometric information to voxel representation via geometry-to-voxel auxiliary mechanism.
Our auxiliary task is to predict point labels for local point patches with point branch and point decoder. Therefore, the final loss function is formulated as:
\begin{equation}
    \mathcal{L}=\mathcal{L}_{v}+\mu\mathcal{L}_{p}, \label{eq:loss}
\end{equation}
where $\mathcal{L}_{v}$ and $\mathcal{L}_{p}$ stand for per-voxel and per-point cross entropy loss and $\mu$ is the weight of the point loss.

\subsection{Discussion}
In this section, we highlight previous point-voxel networks~\cite{PVCNN2019NIPS, SPVCNN2020ECCV, PVT2021IJIS} which combine a point-based branch and a voxel-based branch together to leverage the advantages of both efficient voxel representations and elaborate point features. Additionally, we compare our GeoAuxNet with these methods.
PVCNN~\cite{PVCNN2019NIPS} and PVT~\cite{PVT2021IJIS} rely on point-based networks to extract high quality representations with time-consuming searching or attention mechanisms, while SPVCNN~\cite{SPVCNN2020ECCV} does not explore detailed local geometric information. Since the interaction between two branches in these methods only contains voxelization and devoxelization operations,  they fail to benefit from both voxel representations and point-level elaborate features.
Our GeoAuxNet preserves hierarchical sensor-aware geometry pools as the bridge to enable voxel representations to absorb point-level local geometric information without using the point network during inference.



\section{Experiments}
We validate our proposed GeoAuxNet across point clouds from multiple sensors. In Section~\ref{sec:semantic segmentation}, we train various methods on joint datasets including S3DIS~\cite{S3DIS2016CVPR}, ScanNet~\cite{ScanNet2017CVPR} and SemanticKITTI~\cite{SemanticKITTI2019ICCV}. In Section~\ref{sec:geometry pool}, we analyze the effectiveness of our sensor-aware hierarchical geometry pools. In Section~\ref{sec:efficiency of GeoAuxNet}, we compare GeoAuxNet with typical point-based and voxel-based methods to demonstrate the efficiency of our designed geometry-to-voxel auxiliary learning. In Section~\ref{sec:ablation study}, we ablate different design choices of our voxel-guided dynamic point network. 

\subsection{Semantic Segmentation}\label{sec:semantic segmentation}
\paragraph{Dataset.} We conduct semantic segmentation experiments on the joint data of three datasets: S3DIS~\cite{S3DIS2016CVPR} and ScanNet~\cite{ScanNet2017CVPR}, which are generated from RGB-D cameras, and SemanticKITTI~\cite{SemanticKITTI2019ICCV}, which is captured by LiDAR. The S3DIS dataset comprises 271 rooms from six areas in three distinct buildings, which are densely sampled on the surfaces of the meshes and annotated into 13 categories. Model performance evaluation is typically done using results from Area 5. The ScanNet dataset comprises 1,613 scene scans reconstructed from RGB-D frames. It is divided into 1,201 scenes for training, 312 scenes for validation, and 100 scenes for benchmark testing. SemanticKITTI consists of 22 point cloud sequences, where sequences 00 to 10, 08 and 11 to 21 are used for training, validation and testing. After merging classes with distinct moving status and discarding classes with very few points, a total of 19 classes are selected for training and evaluation.

\paragraph{Implementation Details.}
Our voxel-based backbone is built on MinkowskiNet~\cite{SparseUNet2019CVPR} with the batch normalization operations proposed by PPT~\cite{PPT2023ArXiv}. The voxel size is set to 0.05m.
We train our methods as well as MinkowskiNet SPVCNN~\cite{SPVCNN2020ECCV} and PPT~\cite{PPT2023ArXiv} on joint input data~\cite{S3DIS2016CVPR, ScanNet2017CVPR, SemanticKITTI2019ICCV} captured by RGB-D cameras and LiDAR collectively. We also fine tune our method on three datasets separately. For the sake of fairness, we set the number of parameters of MinkowskiNet, SPVCNN, PPT and GeoAuxNet to about 60M.
We use SGD optimizer and OneCycleLR~\cite{OneCycleLR2017ArXiv} scheduler, with learning rate 0.05 and weight decay $10^{-4}$.
We adopt OneCycleLR~\cite{OneCycleLR2017ArXiv} scheduler with 5\% percentage of the cycle spent increasing the learning rate and a cosine annealing strategy.
We train all models with data augmentation used in PPT~\cite{PPT2023ArXiv}. We set $\lambda$ in Eq.~\ref{eq:absorb} to 0.1, $\mu$ in Eq.~\ref{eq:loss} to 0.1, threshold for geometry pool update $\epsilon$ to 0.9. The total number of iterations is equal to the sum of necessary iterations for all datasets. The sampling ratio across S3DIS, ScanNet and SemanticKITTI is set to 2:2:5. The number of points for the auxiliary branch does not exceed $2\times10^{4}$. The max size of hierarchical geometry pools for each stage is set to 32, 64, 128, 256 separately.

\paragraph{Results Comparison.}
As shown in Table~\ref{tab:main results}, GeoAuxNet surpasses other methods which are trained on three datasets jointly. Our methods outperforms SPVCNN~\cite{SPVCNN2020ECCV}, which does not utilize its point branch to extract local information, by more than 10\% in mIoU on each dataset. PPT~\cite{PPT2023ArXiv} introduces different datasets as prompts to collaboratively train a single model on multiple datasets from the same sensor type.  The language-guided alignment of PPT suffers from the large domain gap caused by different sensors, which fails to converge on the joint training datasets. Therefore, we replace it by the decoupled alignment with separate heads for each dataset, which is also defined in PPT~\cite{PPT2023ArXiv}. We construct sensor-aware geometry pools to address this issue and outperform PPT by about 2\% in mIoU on each dataset. Besides, we also compare with experts trained on each single dataset. We achieve competitive results with point-based methods on datasets from RGB-D cameras. More specifically, we even outperform PointTransformer in validation mIoU of ScanNet. Despite training on inter-domain datasets from different sensors, GeoAuxNet has a higher accuracy on SemanticKITTI than SPVCNN~\cite{SPVCNN2020ECCV}, and attain encouraging performances even compared with recent state-of-the-art expert~\cite{SphereFormer2023CVPR}.
Although we do not introduce any specific designs such as point-wise attention mechanism~\cite{PointTransformer2021ICCV, StratifiedTransformer2022CVPR} and cylinder coordinate system~\cite{Cylinder3D2021CVPR} for sparse LiDAR point clouds,
the quantitative results demonstrate the effectiveness of geometry-to-voxel auxiliary learning by constructing hierarchical geometry pools to provide auxiliary sensor-aware point-level geometric priors for voxel representations.
\begin{figure*}[t]   
  \centering
  \subfloat[]
  {
      \label{fig:gray plane}\includegraphics[height=0.13\textwidth]{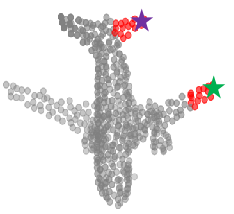}
  }
  \subfloat[]
  {
      \label{fig:colorful plane}\includegraphics[height=0.13\textwidth]{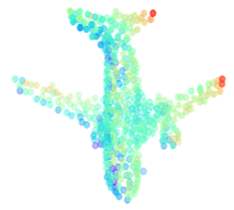}
  }
  \subfloat[]
  {
      \label{fig:gray chair}\includegraphics[height=0.13\textwidth]{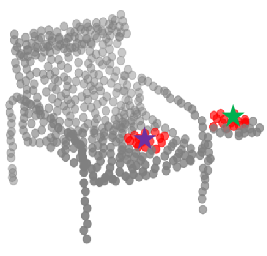}
  }
  \subfloat[]
  {
      \label{fig:colorful chair}\includegraphics[height=0.13\textwidth]{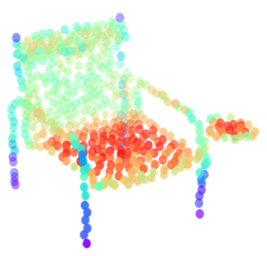}
  }
  \subfloat[]
  {
      \label{fig:gray chair2}\includegraphics[height=0.13\textwidth]{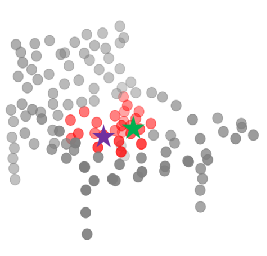}
  }
  \subfloat[]
  {
      \label{fig:colorful chair2}\includegraphics[height=0.13\textwidth]{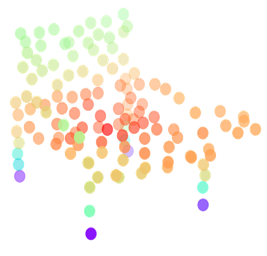}
  }
  \subfloat
  {
      \label{fig:colorbar}\includegraphics[height=0.13\textwidth]{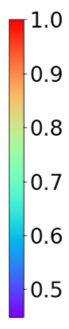}
  }
  \caption{Visualization of the cosine similarity between features. The \textcolor[RGB]{112,48,160}{purple} star is the selected point, and the \textcolor[RGB]{0,176,80}{green} star is the point with a significantly similar feature to the \textcolor[RGB]{112,48,160}{purple} star. We calculate the cosine similarity between the feature of the red star and other features and visualize them in image (b), (d) and (f). The nearest neighbors of the \textcolor[RGB]{112,48,160}{purple} and \textcolor[RGB]{0,176,80}{green} stars are marked \textcolor{red}{red} in image (a), (c) and (e). (a), (b), (c) and (e) are generated from the first stage of the point network and (e), (f) are from the second stage.} 
  \label{fig:robustness} 
\end{figure*}

\subsection{Hierarchical Geometry Pools}\label{sec:geometry pool}
In this section, we further conduct experiments on our hierarchical geometry pools. First, we analyze the similarity of geometric structures in 3D space. Second, we discuss the appropriation to represent geometric information with our geometry pools. Finally, we study the construction of sensor-aware pools for different sensors.

\paragraph{Similarity of Geometric Structures.}
Local geometric structures possess similarity in 3D space. For example, planar structures exist in desks, chairs and airplanes. These spatial structures are not unrelated but consistent. To further analyze the similarity of geometric structures learned by networks, we train our point encoder on ModelNet40~\cite{ShapeNet2015CVPR}. For each point, we use kNN to search 16 nearest neighbors and aggregate them via a max-pooling function to generate local geometric features. We select a center point marked as a purple star in Figure~\ref{fig:gray plane} and Figure~\ref{fig:gray chair}, and calculate the cosine similarity between its features and other point features. Specifically, the wing and the tail of the plane in Figure~\ref{fig:gray plane} have similar local structures, resulting in high similarity of features in Figure~\ref{fig:colorful plane}. Figure~\ref{fig:colorful chair} and Figure~\ref{fig:colorful chair2} show the same phenomenon at different stages in the network, where planar structures are similar to each other, while different from the chair legs. 
Since features from deeper layers possess larger receptive field sizes due to grouping and aggregating operations and comprise more geometric information, our hierarchical pools contain more geometric features for deeper layers.

\begin{figure}[t]
    \centering
    \begin{subfigure}[b]{0.23\textwidth}
        \includegraphics[width=\textwidth]{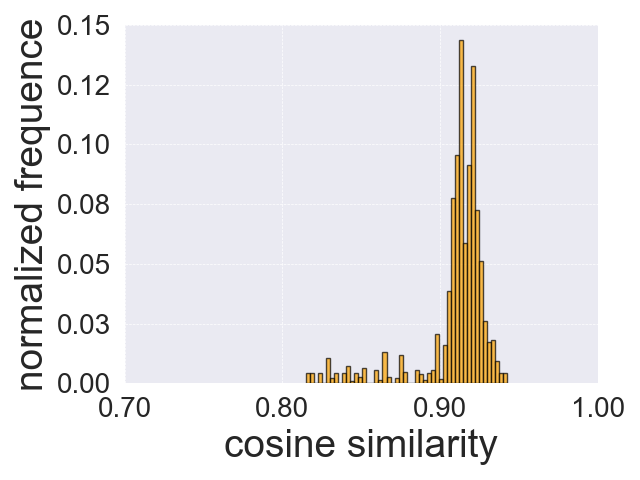}
        \caption{Stage One}
        \label{fig:image1}
    \end{subfigure}
    \hfill 
    \begin{subfigure}[b]{0.23\textwidth}
        \includegraphics[width=\textwidth]{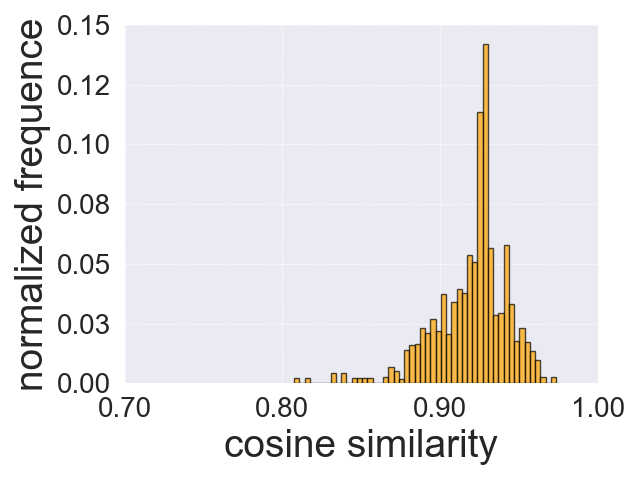}
        \caption{Stage Two}
        \label{fig:image2}
    \end{subfigure}
    
    \begin{subfigure}[b]{0.23\textwidth}
        \includegraphics[width=\textwidth]{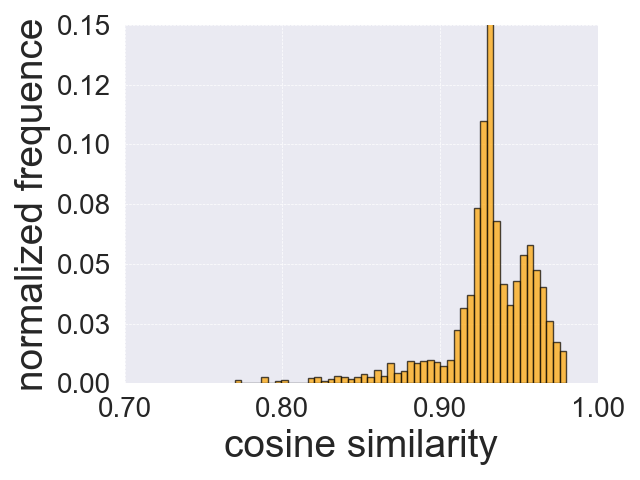}
        \caption{Stage Three}
        \label{fig:image3}
    \end{subfigure}
    \hfill 
    \begin{subfigure}[b]{0.23\textwidth}
        \includegraphics[width=\textwidth]{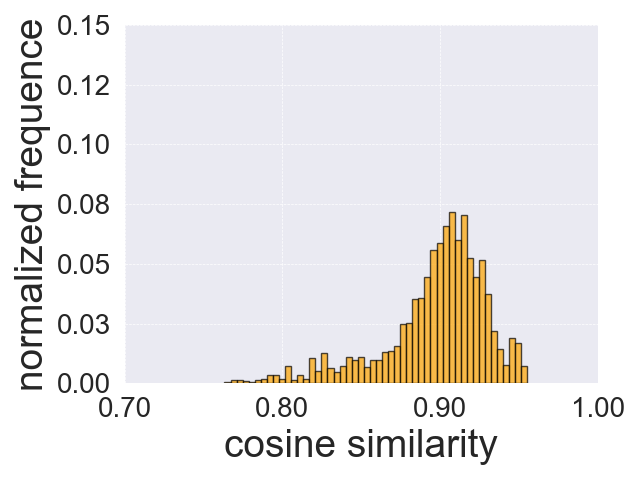}
        \caption{Stage Four}
        \label{fig:image4}
    \end{subfigure}
    
    \caption{Statistical results of cosine similarity between hierarchical geometry pools and point-level features extracted by the point network from point clouds in S3DIS~\cite{S3DIS2016CVPR} at different stages.}
    \label{fig:cosine similarity}
    \vspace{-0.5cm}
\end{figure}

\paragraph{Representativeness of Geometry Pools.}
To demonstrate the representativeness of our pools, we calculate cosine similarity between all features in a scene from S3DIS~\cite{S3DIS2016CVPR} extracted by the point encoder and the features stored in the pool.
For each stage, given the geometry pool $\mathcal{F}^{D}=\{f^{D}_{i}, 1\leq i\leq N\}$ and all features in a scene $\mathcal{F}^{S}=\{f^{S}_{i}, 1\leq i\leq M\}$, the similarity $s_{i}$ between $f^{S}_{i}$ and $\mathcal{F}^{D}$ is formulated as:
\begin{equation}
    s_{i}= \mathop{max}\limits_{1\leq j\leq N}\dfrac{f^{S}_{i}\cdot f^{D}_{j}}{\Vert f^{S}_{i}\Vert \Vert f^{D}_{j}\Vert},
\end{equation}
where $\Vert \cdot \Vert$ stands for L2 norm. 
The distribution of similarities is shown in Figure~\ref{fig:cosine similarity}.
From the statistical data, most features have cosine similarity higher than 0.85, which indicates that each feature can find a similar feature among our learned representatives in the pool.  
In this manner, we avoid using point-based network to process large scale points to generate fine-grained features for each voxel.
Our hierarchical pools contain typical geometric features of scenes.
Therefore, we can fuse point-level geometry information stored in our pools into voxel representations via geometry-to-voxel auxiliary mechanism efficiently.

\begin{figure}[t]
    \centering
    \subfloat[]
      {
          \label{fig:subfig1 similarity}\includegraphics[height=0.43\linewidth]{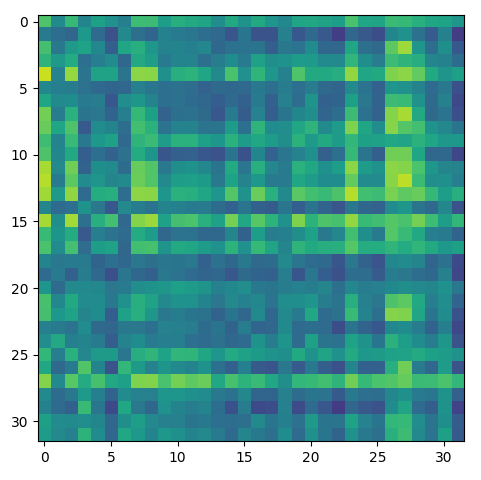}
      }
      \hfill 
      \subfloat[]
      {
          \label{fig:subfig2 similarity}\includegraphics[height=0.43\linewidth]{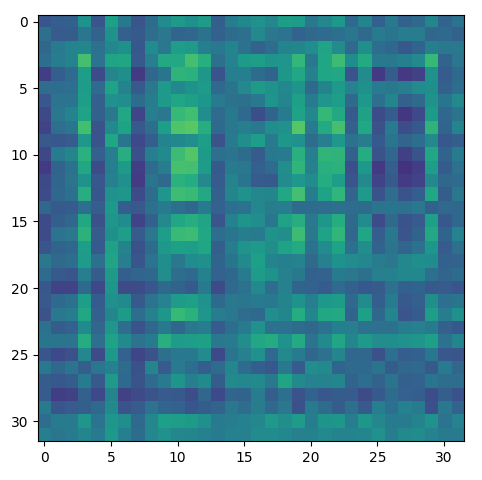}
      }
      \subfloat
      {
          \label{fig:subfig3 similarity}\includegraphics[height=0.43\linewidth]{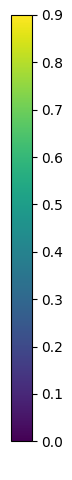}
      }
    \caption{Cosine similarity between geometry pools. (a) shows the similarity between geometry pools of S3DIS~\cite{S3DIS2016CVPR} and ScanNet~\cite{ScanNet2017CVPR}. (b) shows the similarity between geometry pools of S3DIS~\cite{S3DIS2016CVPR} and SemanticKITTI~\cite{SemanticKITTI2019ICCV}. Intra-sensor geometry pools in (a) have higher similarity than inter-sensor geometry pools in (b).}
    \label{fig:sensor-aware}
    \vspace{-0.5cm}
\end{figure}

\paragraph{Sensor-aware Pools for Multiple Sensors.}
Motivated by the diversity of density and sampling patterns of point clouds from various sensors, we construct different geometry pools to preserve relevant geometric information for diverse sensors.
We investigate another two strategies: (i) shared geometry pools across sensors and (ii) different geometry pools for different datasets. We calculate the cosine similarity between features from different pools. As shown in Figure~\ref{fig:sensor-aware}, geometry pools from the same sensor possess higher similarity than those from different sensors. As shown in Table~\ref{tab:sensor-aware}, training with shared geometry pools across sensors leads to a significant decrease on performance due to the inherent diversity of geometric structures from different sensors, while sharing geometry pools with datasets from the same sensor serves as data augmentation to improve the effectiveness of the geometry pools.

\begin{table}[]
    \centering
    \caption{Ablation study of sensor-aware geometry pools. ``Shared intra-sensor" means the geometry pools which are shared for point clouds from the same sensor, while different for point clouds from different sensors. ``Shared inter-sensor" means the geometry pools shared for different sensors. ``Shared intra-dataset" means the geometry pools are different for different datasets. We report mIoU and mAcc on Area 5 of S3DIS~\cite{S3DIS2016CVPR}.}
    \begin{tabular}{l|ll}
    \toprule
        Methods                  & mIoU (\%)            & mAcc (\%) \\ \midrule
        Shared intra-sensor      & 69.5             & 74.5     \\
        Shared inter-sensor      & $61.4_{\textcolor{Red}{(-8.1)}}$  & $68.3_{\textcolor{Red}{(-6.2)}}$ \\
        Shared intra-dataset     & $67.5_{\textcolor{Red}{(-2.0)}}$  & $73.3_{\textcolor{Red}{(-1.2)}}$ \\
        \bottomrule
    \end{tabular}
    \label{tab:sensor-aware}
\end{table}

\begin{table}[]
\centering
\caption{Efficiency analysis. We report the inference time and throughput on pre-processed data from S3DIS dataset.}
\begin{tabular}{lc|cc}
\toprule
Method & Params. & \begin{tabular}[c]{@{}c@{}}Inference \\ Time (ms)\end{tabular} & \begin{tabular}[c]{@{}c@{}}Throughput\\ (ins./sec.)\end{tabular} \\ \midrule
PointNet++~\cite{PointNet++2017NIPS}   & 1.0M       & 378.7               & 108 \\
PT~\cite{PointTransformer2021ICCV}     & 7.8M        & 1079.7              & 34  \\
PTv2~\cite{PointTransformerV22022NIPS} & 3.9M        & 4732.2              & 12  \\
MinkowskiNet~\cite{SparseUNet2019CVPR} & 60.9M       & 45.4                & 898 \\
SPVCNN~\cite{SPVCNN2020ECCV}           & 61.0M       & 48.4                & 682 \\
PPT~\cite{PPT2023ArXiv}                & 63.0M       & 80.4                & 251 \\
GeoAuxNet                              & 64.7M       & 51.8                & 572 \\
       \bottomrule
\end{tabular}
\label{tab:efficiency}
\vspace{-0.5cm}
\end{table}
\subsection{Efficiency of GeoAuxNet}\label{sec:efficiency of GeoAuxNet}
We validate the efficiency of GeoAuxNet compared with other typical point-based~\cite{PointTransformer2021ICCV, PointTransformerV22022NIPS} and voxel-based methods~\cite{SparseUNet2019CVPR, PPT2023ArXiv}. We focus on the mean inference time and throughput by employing a standardized pre-processed point cloud with instance labels from S3DIS Area 5 dataset as the input on a single NVIDIA A40 GPU. 
 Initially, the device undergoes a ``warm-up" phase, where the model is fed with the input point cloud ten times consecutively to eliminate any latency associated with data loading. Subsequently, the model is subjected to an additional 300 iterations of input feeding to calculate mean inference time. To effectively quantify the throughput, we measure the number of instances that a model processes per second. This systematic approach ensures a thorough and reliable assessment of our GeoAuxNet and other state-of-the-art methods.

 In Table~\ref{tab:efficiency}, we show the number of parameters, inference time and throughput for typical point-based and voxel-based networks. Specifically, our GeoAuxNet outperforms the expert model of PointTransformer~\cite{PointTransformer2021ICCV} which is trained on ScanNet~\cite{ScanNet2017CVPR} individually in mIoU with $16.8\times$ throughput and $20.8\times$ inference speedup. 
 Even compared with the simple PointNet++~\cite{PointNet++2017NIPS} whose parameter size is $1/68$ of ours, GeoAuxNet has less inference time and higher throughput. 
 Besides, our method has similar inference time and throughput to voxel-based methods but achieve better performance according to the results in Table~\ref{tab:main results}. GeoAuxNet surpasses PPT~\cite{PPT2023ArXiv} on each benchmark about 2\% in mIoU with $2.3\times$ throughput and $1.6\times$ inference speedup. Our method maintains efficiency for large scale point clouds, while achieves encouraging performances on datasets from multiple sensors.

\begin{table}[]
    \centering
    \caption{Ablation study of our voxel-guided hypernetwork. We report the mIoU and mAcc on Area 5 of S3DIS~\cite{S3DIS2016CVPR}.}
    \begin{tabular}{l|ll}
    \toprule
        Methods                  & mIoU (\%)        & mAcc (\%) \\ \midrule
        Vanilla                  & 63.3             & 68.7     \\
        $+$ voxel-guided     & $66.0_{\textcolor{Green}{(+2.7)}}$  & $72.5_{\textcolor{Green}{(+3.8)}}$ \\
        $+$ relative positions   & $66.8_{\textcolor{Green}{(+0.8)}}$  & $73.2_{\textcolor{Green}{(+0.7)}}$ \\
        $+$ stage latent code    & $67.6_{\textcolor{Green}{(+0.8)}}$  & $73.3_{\textcolor{Green}{(+0.1)}}$ \\
        \bottomrule
    \end{tabular}
    \label{tab:ablation hypernet}
    \vspace{-0.5cm}
\end{table}

\subsection{Voxel-guided Hypernetworks}\label{sec:ablation study}

To investigate the architecture design of our voxel-guided hypernetwork,  we conduct ablation studies on S3DIS semantic segmentation. 
We first introduce a vanilla point network which learns the weight and bias directly. Then, we adopt a hypernetwork taking voxel features as input to generate weights and biases for the point network. We further take account of the relative positions of points. Finally, we optimize learnable stage latent codes to instruct the information of different layers. We train these four models on S3DIS dataset. As shown in Table~\ref{tab:ablation hypernet}, the exploration of voxel features provides significant guidance for the point network to learn high quality geometric information, leading to an increase in test accuracy by 2.7\% in mIoU and 3.8\% in mAcc. Relative positions and stage latent codes also improve the performance of our methods.

\section{Conclusion}

In this paper, we propose GeoAuxNet for multi-sensor point clouds with designed geometry-to-voxel auxiliary learning. We construct hierarchical geometry pools to learn auxiliary sensor-aware point-level geometric priors at different layers for sensor-agnostic voxel features. To generate our geometry pools, we also introduce a voxel-guided dynamic point network to leverage voxel prior knowledge for elaborate point features extraction.  Our proposed geometry-to-voxel auxiliary builds a bridge between point-level and voxel-level features in an efficient manner without using the point network during inference time. Experimental results have shown the effectiveness and efficiency of our methods. We hope this work will inspire future research towards universal representation learning for point clouds. 

\paragraph{Acknowledgements.} This work was supported by the National Natural Science Foundation of China under  Grant 62206147.




{
    \small
    \bibliographystyle{ieeenat_fullname}
    \bibliography{main}
}

\end{document}